\documentclass[10pt,twocolumn,letterpaper]{article}

\usepackage[pagenumbers]{wacv}

\usepackage{stfloats}
\usepackage{placeins}
\usepackage{multirow}
\usepackage{makecell}
\usepackage{colortbl}
\usepackage{array}
\usepackage[normalem]{ulem}
\usepackage{bm}
\usepackage{bbm}
\usepackage{tikz}
\usepackage{algorithm}
\usepackage{algorithmicx}
\usepackage{algpseudocode}

\definecolor{skyblue}{HTML}{DEEBFE}
\definecolor{rankone}{HTML}{FFC1C3}
\definecolor{ranktwo}{HTML}{FFC991}
\definecolor{rankthree}{HTML}{FFF6A9}

\newcommand{\rankone}[1]{\cellcolor{rankone}#1}
\newcommand{\ranktwo}[1]{\cellcolor{ranktwo}#1}
\newcommand{\rankthree}[1]{\cellcolor{rankthree}#1}
\newcommand{\rankbox}[2]{%
  \leavevmode%
  \begingroup\setlength{\fboxsep}{1pt}%
  \colorbox{#1}{\hspace*{2pt}\vphantom{Ay}#2\hspace*{2pt}}%
  \endgroup
}

\newcommand{\abbr}{\textsc{ClunoGS}}
\newcommand{\assign}{$\leftarrow$}
\newcommand*\Circled[2][gray!40]{%
  \tikz[baseline=(char.base)]{\node[
    shape=circle,
    draw=none,
    thick,
    fill=#1,
    inner sep=0.9pt
  ] (char) {\textcolor{black}{#2}};}
}

\definecolor{wacvblue}{rgb}{0.21,0.49,0.74}
\usepackage[breaklinks,colorlinks,allcolors=wacvblue]{hyperref}

\title{\abbr: Compact Language 3D Gaussian Splatting via Unified Optimization}

\author{%
Jian Xu\quad Yi Li\quad Ying Huang\quad Siyi Wu\\[3pt]
Bingzhe Li\quad Yuede Ji\quad Miao Yin\\[6pt]
{\tt\small \{jian.xu,ying.huang,siyi.wu,yuede.ji,miao.yin\}@uta.edu}\\[2pt]
{\tt\small \{yi.li3,bingzhe.li\}@utdallas.edu}%
}

\hypersetup{
    pdftitle={ClunoGS: Compact Language 3D Gaussian Splatting via Unified Optimization},
    pdfauthor={Jian Xu, Yi Li, Ying Huang, Siyi Wu, Bingzhe Li, Yuede Ji, Miao Yin}
}

\begin{document}
\twocolumn[{
    \maketitle
    \centering
    \begin{minipage}{1.00\textwidth}
        \centering
        \includegraphics[width=\textwidth]{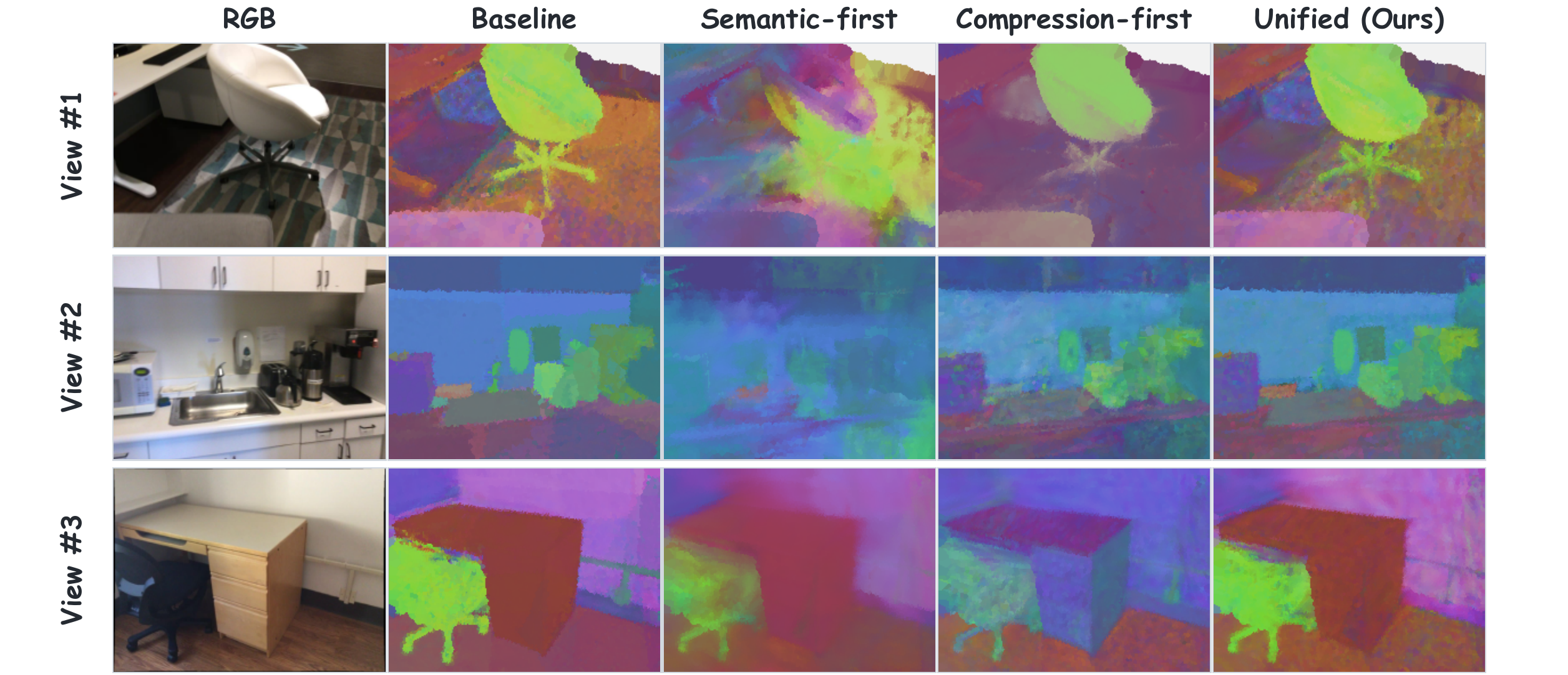}
        \vspace{0.1cm}
    \end{minipage}
    \vspace{-2.5mm}
    \captionsetup{hypcap=false}
    \captionof{figure}{
    Impact of pipeline ordering in semantic 3DGS compression. Semantic-first
    learns semantics before compression, whereas Compression-first compresses
    the 3DGS before semantic learning. PCA visualizations show that both
    sequential pipelines degrade semantic representations, while our unified
    pipeline better preserves them.}
    \label{fig:teaser}
    \vspace{0.3cm}
}]

\begin{abstract}
Language 3D Gaussian Splatting enables open-vocabulary scene understanding by
associating Gaussian primitives with language features. These semantic
attributes increase model storage, while most existing 3DGS compression
methods are designed mainly to preserve RGB rendering quality. One
straightforward approach is to apply sparsification and quantization
sequentially. However, pruning based only on rendering quality can remove
Gaussians that have small RGB contributions but are important for semantic
prediction. Once removed, their semantic information cannot be recovered by
later quantization. We propose \abbr{}, a compact framework for Language 3DGS
that combines semantics-aware Gaussian sparsification and vector quantization
in a unified constrained objective. Semantic supervision is used during
sparsification, and the objective jointly considers rendering quality,
semantic accuracy, and compactness. We optimize the unified objective with
alternating primal, sparsification, quantization, and dual updates.
We evaluate \abbr{} on open-vocabulary localization and segmentation using the
LERF and ScanNet-10 benchmarks. Under aggressive compression, our unified,
semantics-aware formulation substantially alleviates the degradation of
semantic representations caused by sequential compression and
semantics-agnostic Gaussian sparsification. Ablations further show that
jointly coordinating sparsification and quantization better preserves
semantic accuracy than applying them independently.
\end{abstract}
    
\section{Introduction}
\label{sec:intro}

Open-vocabulary 3D scene understanding enables objects in a 3D scene to be
localized or segmented using arbitrary natural-language queries, without
restricting predictions to a predefined class set~\citep{peng2023openscene}.
This capability supports applications such as
AR/VR~\citep{koch2024lang3dsg}, autonomous
driving~\citep{cheng2024open}, and robotic
manipulation~\citep{qiu2024open,huang2022visual}. Earlier neural-field methods
encoded semantics in implicit representations, but repeated volumetric
sampling and network evaluation made training and rendering
costly~\citep{wang2023semantic,kerr2023lerf}. Language 3DGS instead attaches
vision-language features to explicit Gaussian primitives and reuses the
tile-based splatting pipeline of 3DGS to render semantic feature maps
efficiently, avoiding repeated volumetric sampling and per-sample network
evaluation~\citep{kerbl20233d,zhou2024feature,qin2024langsplat}.

Directly attaching high-dimensional vision-language features to individual
Gaussians incurs substantial storage and rendering
overhead~\citep{zhou2024feature,qin2024langsplat}. LangSplat mitigates this
problem using a scene-specific low-dimensional autoencoder, while LangSplatV2
further replaces the heavyweight decoder with sparse per-Gaussian
coefficients over shared dictionaries, enabling efficient high-dimensional
feature splatting~\citep{qin2024langsplat,li2025langsplatv2}. These advances
reduce the cost of semantic feature representation and inference, but do not
target compression of the underlying RGB Gaussian scene. Consequently, a
complete Language 3DGS model still contains the original 3DGS attributes
together with additional semantic parameters. Its compression must therefore
preserve both photometric fidelity and semantic responses.

Standard 3DGS compression often combines sparsification or
pruning~\citep{yang2024spectrally,ali2024trimming,zhang2025gaussianspa,
hanson2025pup,fan2024lightgaussian}
with
quantization~\citep{navaneet2023compact3d,lee2024compact,navaneet2024compgs}.
Applying them sequentially decouples semantic learning from Gaussian
selection, so RGB-driven sparsification can discard semantically important
Gaussians that subsequent stages cannot recover. Figure~\ref{fig:teaser}
illustrates this failure using LangSplatV2~\citep{li2025langsplatv2},
GaussianSpa, and CompGS: semantic representations degrade whether semantic
learning occurs before or after RGB-driven Gaussian sparsification. Our
unified formulation instead coordinates semantic learning, sparsification,
and quantization throughout optimization.

To address this gap, we propose \abbr{} (\textbf{C}ompact \textbf{L}anguage
via \textbf{Un}ified \textbf{O}ptimization for 3D \textbf{G}aussian
\textbf{S}platting). The framework combines RGB reconstruction and semantic
supervision with a cardinality constraint on Gaussian opacities and vector
quantization (VQ) of selected attributes. Updating the scene parameters, sparse
support, and quantized representations within the same constrained problem
allows them to adapt to one another instead of inheriting irreversible
decisions from a preceding stage. We solve the resulting non-smooth problem
with an alternating primal--dual procedure.

Our contributions are:

\begin{itemize}[leftmargin=*,topsep=0pt,itemsep=3pt,parsep=0pt]
\item We introduce a unified constrained formulation that couples an
RGB--semantic task objective with Gaussian sparsification and attribute vector
quantization for compact Language 3DGS.

\item We develop a compact Language 3DGS optimization framework and solve the
unified problem using an alternating primal--dual procedure that coordinates
differentiable scene optimization, sparsification, quantization, and dual
updates.

\item We evaluate \abbr{} on LERF and ScanNet-10 using recent language-field
baselines, sequential compression and ordering controls, qualitative
comparisons, and controlled ablations. Across the two benchmarks, \abbr{}
reduces serialized storage by 80.9--82.8\% relative to the corresponding
uncompressed models, with losses of at most 1.6 mIoU points, 1.1 accuracy
points, and 0.07~dB PSNR.

\end{itemize}

\section{Related Work}
\label{sec:related}
\paragraph{Language 3DGS.}
Language-aware 3D representations lift 2D foundation-model features into
queryable 3D fields, with LERF providing an early NeRF-based
formulation~\citep{kerr2023lerf}. Language 3DGS methods mainly follow two
paradigms. Render-supervised methods learn per-Gaussian features or codes by
matching rendered feature maps to 2D supervision, as in LangSplat, Feature
3DGS, GAGS, and LangSplatV2~\citep{qin2024langsplat,zhou2024feature,
peng2026gags,li2025langsplatv2}. Direct-lifting methods instead register or
aggregate multi-view features onto reconstructed Gaussians without iterative
semantic fitting, as in Dr. Splat and Occam's
LGS~\citep{jun2025dr,cheng2025occam}. LightSplat and
ExtrinSplat further replace dense per-Gaussian embeddings with compact semantic
indices or an external object-group index~\citep{bang2026lightsplat,
ding2026extrinsplat}. We build on LangSplatV2 because its semantic coefficients
and RGB attributes share the same Gaussian support, allowing sparsification and
attribute quantization to account for both rendering and semantic quality.

\paragraph{Compact Language 3DGS.}
Recent work has explored compact representations for the semantic features in
Language 3DGS. DF-3DGS separates RGB and semantic Gaussians and sparsifies the
semantic field, while CF3 compresses feature Gaussians through pruning,
merging, and autoencoding~\citep{dai2025df3dgs,lee2025cf3}.
CSGaussian progressively entropy-codes both color and semantic
attributes~\citep{tseng2026csgaussian}. SCOUP represents semantic features
with sparse codes, whereas CoSAG predicts and entropy-codes compact semantic
bindings for individual Gaussians~\citep{budimir2026scoup,jia2026cosag}.
These methods reduce semantic storage through specialized feature
representations and coding schemes.

\paragraph{General 3DGS Compression.}
RGB-only 3DGS compression reduces active primitives, attribute storage, or
bitstream size. GaussianSpa and PUP 3D-GS sparsify optimized scenes using
constrained or uncertainty-based
criteria~\citep{zhang2025gaussianspa,hanson2025pup}. Compact 3DGS, C3DGS, and
CompGS use clustering or quantization to reduce attribute
storage~\citep{lee2024compact,niedermayr2024compressed,navaneet2024compgs}.
RDO-Gaussian jointly optimizes masks and entropy-constrained attributes, while
HAC++ and LocoGS model context or local coherence for compact
coding~\citep{wang2024end,chen2025hacpp,shin2025locogs}. FCGS applies a
feed-forward codec to an optimized Gaussian scene~\citep{chen2025fcgs}. Joint
sparsification and quantization are therefore established in RGB 3DGS. The
concurrent GETA-3DGS preprint also studies joint structured pruning and
heterogeneous quantization~\citep{zhang2026geta}. \abbr{} instead optimizes
opacity with both RGB and semantic supervision before cardinality projection,
and coordinates Gaussian sparsification with attribute vector quantization in
the same constrained formulation.

\section{Background}
3D Gaussian Splatting uses a collection of Gaussians to model
3D scene geometry and appearance. Specifically, each Gaussian $G(\bm{x})$ is
defined by a mean vector $\bm{\mu} \in \mathbb{R}^3$ and a covariance matrix
$\bm{\Sigma}$:
\begin{equation}
    G(\bm{x}) = \exp \Big(- \frac{1}{2} (\bm{x}-\bm{\mu})^{\top}
    \bm{\Sigma}^{-1} (\bm{x} -  \bm{\mu}) \Big),
    \label{eq:3dGaussian_refined}
\end{equation}
where $\bm{x}$ is a location in the 3D scene. The learnable parameters of a
Gaussian are $\{\bm{\mu},\bm{c},o,\bm{R},\bm{S}\}$, which correspond to its
position, color, opacity, rotation, and scale.

To optimize the Gaussian parameters, they are projected onto 2D image planes,
and tile-based rasterization~\citep{zwicker2001ewa} is used for efficiency.
The pixel color $\bm{C}_\text{pixel}$ is rendered by:
\begin{equation}
    \bm{C}_\text{pixel}
    = \sum_{i \in \mathcal{N}} \bm{c}_i \, \alpha_i
    \prod_{j=1}^{i-1} (1 - \alpha_j),
    \quad
    \alpha_i = o_i \, G_i',
    \label{eq:rendering_3dgs_refined}
\end{equation}
where $\bm{c}_i$ is the color of the $i$-th Gaussian, $o_i$ is its opacity,
$\mathcal{N}$ is the set of ordered Gaussians contributing to the
rasterization at the target rendering pixel, and $G_i'$ represents the
projection of the Gaussian onto the 2D plane.

LangSplatV2~\citep{li2025langsplatv2} attaches sparse semantic coefficients to
each Gaussian. For semantic level $h\in\{1,2,3\}$, let
$\bm{w}_i^{(h)}\in\mathbb{R}^{1\times L}$ denote the coefficient vector of
Gaussian $i$, and let $\bm{B}^{(h)}\in\mathbb{R}^{L\times D}$ denote the
scene-level semantic dictionary shared by all Gaussians at that level. Only
the top-$K$ entries of $\bm{w}_i^{(h)}$ are nonzero, and these entries are
normalized to sum to one. The high-dimensional feature is reconstructed after
coefficient rendering and is not stored as a separate per-Gaussian vector.

The sparse coefficients can be rendered before feature reconstruction. For
pixel $p$, let $\alpha_{i,p}=o_iG_i'(p)$ and
$e_{i,p}=\alpha_{i,p}\prod_{j<i}(1-\alpha_{j,p})$. The rendered semantic
feature is
\begin{equation}
    \bm{F}_p^{(h)}=
    \left(\sum_{i\in\mathcal{N}_p}e_{i,p}\bm{w}_i^{(h)}\right)\bm{B}^{(h)}.
    \label{eq:rendering_lang_refined}
\end{equation}
This rendering shares the geometry and opacity-dependent compositing weights
of RGB rendering. The rendered features are compared with text queries to
produce per-pixel relevance maps for localization or segmentation under the
corresponding benchmark protocol~\citep{kerr2023lerf}.

\section{Methodology: \abbr}
\label{sec4}

To compact a Language 3DGS model while preserving semantic and RGB quality,
we propose a unified compression framework, \abbr{}. Our method places
semantics-aware Gaussian sparsification and vector quantization in a single
constrained objective. We solve it with a primal--dual procedure that
alternates among differentiable scene optimization, sparsification,
quantization, and dual updates.

\begin{figure*}[!t]
    \centering
    \includegraphics[width=\textwidth]{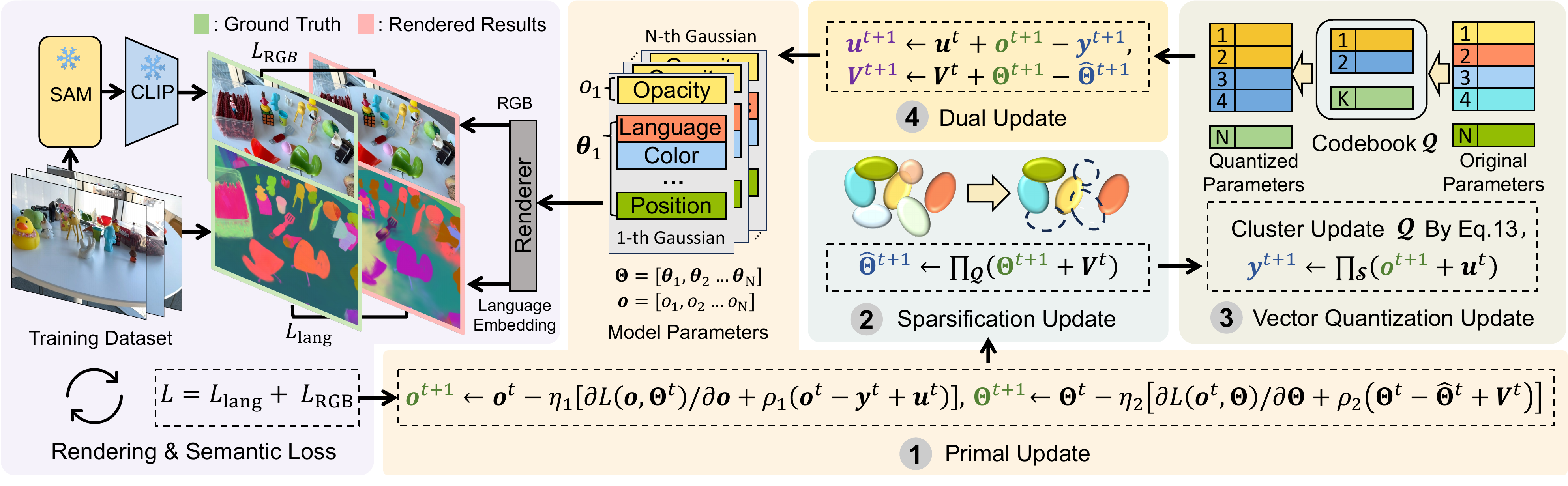}
    \caption{Overview of the \abbr~framework, illustrating the four updates
    performed in each iteration of the optimization loop. The diagram
    suppresses the language-loss weight and attribute-group indices.}
    \label{fig:overall}
\end{figure*}

\subsection{Problem Formulation}
\label{subsec:problem_formulation}

\paragraph{Language-Embedded Training Objective.}
In standard 3DGS training~\citep{kerbl20233d}, the Gaussian parameters are
learned using a combination of pixelwise $\ell_1$ loss and differentiable SSIM
loss between rendered RGB images and ground-truth multi-view images. The
rendering loss is
$L_{\mathrm{RGB}}=(1-\lambda)L_1+\lambda L_{\mathrm{D\text{-}SSIM}}$,
where $\lambda\in[0,1]$ balances the two terms.

For semantic supervision, we follow LangSplatV2. As introduced in the previous
section, each Gaussian is associated with sparse semantic coefficients over
scene-level dictionaries, and the high-dimensional semantic features are
reconstructed after coefficient rendering. We match them to the SAM-derived
CLIP targets $\bar{\bm{F}}_p^{(h)}$ at all three semantic levels using
pixelwise cosine distance:
\begin{equation}
L_{\mathrm{lang}}=\frac{1}{3|\mathcal{P}|}
\sum_{h=1}^{3}\sum_{p\in\mathcal{P}}
\left(1-\frac{\langle\bm{F}_p^{(h)},\bar{\bm{F}}_p^{(h)}\rangle}
{\|\bm{F}_p^{(h)}\|_2\,\|\bar{\bm{F}}_p^{(h)}\|_2}\right),
\label{eq:language_loss}
\end{equation}
where $\mathcal{P}$ is the set of image pixels. The RGB and language objectives
are combined as
\begin{equation}
L = L_{\mathrm{RGB}} + \gamma L_{\mathrm{lang}},
\label{eq:training_loss}
\end{equation}
where $\gamma$ denotes the weighting factor that balances the rendering and
language losses. Since RGB and semantic rendering share the same
opacity-dependent compositing weights, semantic supervision can also affect the
Gaussian opacities and thereby influence which Gaussians are retained during
sparsification.

\paragraph{Unified Objective with Sparsity and Vector Quantization.}
From Eqs.~\ref{eq:rendering_3dgs_refined}
and~\ref{eq:rendering_lang_refined}, Gaussian opacity directly affects both RGB
and semantic compositing. We therefore impose sparsity on opacity to reduce the
number of active Gaussian primitives and use vector quantization to reduce the
storage of the remaining continuous per-Gaussian attributes.

Given a 3DGS model with $N$ initial Gaussians, we represent their opacities as
$\bm{o}=[o_1,o_2,\dots,o_N]\in\mathbb{R}^{N}$, where $o_i$ is the opacity of
the $i$-th Gaussian. We use
$\bm{\Theta}=\{\bm{\theta}_1,\bm{\theta}_2,\dots,\bm{\theta}_N\}$ to denote
the continuous per-Gaussian attributes subject to vector quantization. In our
implementation, these include color coefficients, higher-order spherical
harmonic coefficients, scale, rotation, and semantic coefficient values. The
3D positions and scene-level semantic dictionaries remain in full precision,
while the discrete top-$K$ semantic atom indices are stored separately.

Let $\bm{\mathcal{Q}}=\{(Q_j,\bm{q}_j)\}_{j=1}^{M}$ denote the quantization
state, where $Q_j$ is a cluster and $\bm{q}_j$ is its centroid. The centroids
form the quantization codebook. For concise notation, we suppress
attribute-group indices below; in implementation, vector quantization is
applied separately to different attribute groups using independent codebooks.

With the clusters and codebook defined, the unified optimization objective is
given by
\begin{equation}
\begin{aligned}
\min_{\bm{o}, \bm{\Theta}, \bm{\mathcal{Q}}}\quad
 L(\bm{o}, \bm{\Theta})
  &+ \sum_{j=1}^M \sum_{\bm{\theta}_i\in Q_j}
  \|\bm{\theta}_i-\bm{q}_j\|_2^2, \\
\text{s.t.}&\quad
\mathbf{card}(\bm{o})\leq\kappa.
\label{eq:final_objective}
\end{aligned}
\end{equation}

The \textit{sparsity constraint} $\mathbf{card}(\bm{o})\leq\kappa$ limits the
number of Gaussians with nonzero opacity to at most $\kappa$. Here,
$\mathbf{card}(\bm{o})$ is the number of nonzero elements in $\bm{o}$.
Importantly, the opacity parameters are optimized under both RGB and semantic
supervision before the sparsity constraint is enforced. The retained
Gaussians therefore depend on their contribution to the joint RGB--semantic
objective rather than on RGB reconstruction alone. The same decision is
applied to all point-aligned attributes of each Gaussian.

The \textit{quantization penalty}
$\sum_{j=1}^{M}\sum_{\bm{\theta}_i\in Q_j}
\|\bm{\theta}_i-\bm{q}_j\|_2^2$
penalizes deviations of each parameter vector $\bm{\theta}_i$ from its assigned
cluster centroid $\bm{q}_j$. This encourages the continuous attributes to
approach a small set of representative values. After training, each vector can
be approximated by its cluster centroid, requiring only its codeword index and
the corresponding codebook to be stored.

\subsection{Optimization}
\label{sec:optimization}

Figure~\ref{fig:overall} provides an overview of our four-step primal--dual
optimization. The unified objective in Eq.~\ref{eq:final_objective} contains
two non-differentiable components: the \textit{sparsity constraint}
$\mathbf{card}(\bm{o})\leq\kappa$ and the discrete assignments in vector
quantization. We introduce auxiliary variables to separate differentiable
scene optimization from sparsification and quantization.

We first introduce an auxiliary variable $\bm{y}$ to separate the sparsity
constraint from $\bm{o}$, and use
$\bm{\hat{\Theta}}=\{\bm{\hat{\theta}}_1,\dots,\bm{\hat{\theta}}_N\}$
as the quantized copy of $\bm{\Theta}$. We define $g(\cdot)$ as the indicator
function of the sparsity constraint: $g(\bm{y})=0$ when
$\mathbf{card}(\bm{y})\leq\kappa$ and $g(\bm{y})=+\infty$ otherwise.
Hence, Eq.~\ref{eq:final_objective} can be reformulated as
\begin{equation}
\begin{aligned}
\min_{\bm{o}, \bm{\Theta}, \bm{y}, \bm{\hat{\Theta}}, \bm{\mathcal{Q}}}\;
  L(\bm{o}, \bm{\Theta}) &+ g(\bm{y})
  + \sum_{j=1}^M \sum_{\bm{\hat{\theta}}_i\in Q_j}
  \|\bm{\hat{\theta}}_i-\bm{q}_j\|_2^2, \\[3pt]
\qquad \text{s.t.}\;&
  \bm{o}=\bm{y},\;
  \bm{\Theta}=\bm{\hat{\Theta}}.
\end{aligned}
\label{eq:final_objective_refine}
\end{equation}

This reformulation allows the differentiable variables $\bm{o}$ and
$\bm{\Theta}$ to be optimized using gradient-based methods, while the
sparsity constraint and discrete vector-quantization assignments are handled
through the corresponding auxiliary variables.
The resulting alternating procedure is summarized in Alg.~\ref{alg:alg1}, and
its four updates are derived below.

{
\begin{algorithm}[t!]
\caption{Unified Optimization for \abbr}
\label{alg:alg1}

\textbf{Input:} Language 3DGS variables $\bm{\Theta}$, opacity $\bm{o}$,
Gaussian budget $\kappa$, number of clusters $M$, penalty coefficients
$\rho_1,\rho_2$, learning rates $\eta_1,\eta_2$, maximum iterations $T$,
and convergence thresholds $\epsilon_1,\epsilon_2$. \par
\textbf{Output:} Retained Gaussians, quantized parameters
$\bm{\hat{\Theta}}$, and quantization state $\bm{\mathcal{Q}}$.

\begin{algorithmic}[1]
    \State Initialize $\bm{\mathcal{Q}}$ by $k$-means on $\bm{\Theta}$;
    \State $\bm{y}$ \assign $\bm{o}$,
    $\bm{\hat{\Theta}}$ \assign
    $\Pi_{\bm{\mathcal{Q}}}(\bm{\Theta})$,
    $\bm{u}$ \assign $\mathbf{0}$,
    $\bm{V}$ \assign $\mathbf{0}$;
    \State $r_o,r_\Theta$ \assign $\infty$, \quad $t$ \assign $0$;

    \While{$(r_o > \epsilon_1$ \textbf{or} $r_\Theta > \epsilon_2)$
    \textbf{and} $t < T$}
        \State Update $\bm{o},\bm{\Theta}$ with Eq.~\ref{eq:o_theta_update};
        \Comment{\textit{Primal Update}}

        \State Update $\bm{y}$ with Eq.~\ref{eq:sparse};
        \Comment{\textit{Sparsification}}

        \State Update $\bm{\mathcal{Q}},\bm{\hat{\Theta}}$ with
        Eqs.~\ref{eq:quant_assign}--\ref{eq:admm_projection};
        \Comment{\textit{Vector Quant.}}

        \State Update $\bm{u},\bm{V}$ with
        Eqs.~\ref{eq:dual_u}--\ref{eq:dual_V};
        \Comment{\textit{Dual Update}}

        \State $r_o$ \assign $\|\bm{o}^{t+1}-\bm{y}^{t+1}\|_2$,
        $r_\Theta$ \assign
        $\|\bm{\Theta}^{t+1}-\bm{\hat{\Theta}}^{t+1}\|_2$;
        \State $t$ \assign $t+1$;
    \EndWhile
\end{algorithmic}
\end{algorithm}
}

We solve the equality-constrained problem in
Eq.~\ref{eq:final_objective_refine} by constructing an augmented Lagrangian
with scaled Lagrange multipliers $\bm{u},\bm{V}$ and penalty coefficients
$\rho_1,\rho_2$:
\begin{equation}
\begin{aligned}
L_{\rho}(\bm{o}, \bm{\Theta}, &\bm{y}, \bm{\hat{\Theta}},\bm{u},\bm{V}, \bm{\mathcal{Q}}) = \\
& L(\bm{o}, \bm{\Theta}) + g(\bm{y})
+ \sum_{j=1}^M \sum_{\bm{\hat{\theta}}_i\in Q_j}
\|\bm{\hat{\theta}}_i-\bm{q}_j\|_2^2 \\
&+ \frac{\rho_1}{2}\|\bm{o}-\bm{y}+\bm{u}\|_2^2
+ \frac{\rho_2}{2}\|\bm{\Theta}-\bm{\hat{\Theta}}+\bm{V}\|_2^2.
\end{aligned}
\label{eq:Lagrangian}
\end{equation}

The quadratic penalties enforce the equality constraints.

\paragraph{\Circled{1} Primal Update.}
At iteration $t$, only the model parameters $\bm{o}$ and $\bm{\Theta}$ are
updated; all other variables remain fixed. This gives the differentiable
subproblem:
\begin{equation}
\begin{aligned}
(\bm{o}^{t+1}, \bm{\Theta}^{t+1})
= &\arg\min_{\bm{o}, \bm{\Theta}}
L(\bm{o}, \bm{\Theta})
+ \frac{\rho_1}{2}
\|\bm{o}-\bm{y}^{t}+\bm{u}^{t}\|_2^2 \\
+& \frac{\rho_2}{2}
\|\bm{\Theta}-\bm{\hat{\Theta}}^{t}+\bm{V}^{t}\|_2^2.
\end{aligned}
\label{eq:admm_update_param}
\end{equation}

Here, $L(\bm{o},\bm{\Theta})$ is the differentiable RGB--semantic training
objective, while the two quadratic terms couple the continuous parameters to
their sparse and quantized counterparts. Consequently,
Eq.~\ref{eq:admm_update_param} can be optimized using standard gradient-based
optimizers. The gradients for $\bm{o}$ and $\bm{\Theta}$ at iteration $t$ are
\begin{align}
\left(\frac{\partial L_{\rho}}{\partial \bm{o}}\right)^t
&=
\frac{\partial L(\bm{o},\bm{\Theta}^{t})}{\partial \bm{o}}
+\rho_1
\big(\bm{o}^{t}-\bm{y}^{t}+\bm{u}^{t}\big), \\
\left(\frac{\partial L_{\rho}}{\partial \bm{\Theta}}\right)^t
&=
\frac{\partial L(\bm{o}^{t},\bm{\Theta})}{\partial \bm{\Theta}}
+\rho_2
\big(\bm{\Theta}^{t}-\bm{\hat{\Theta}}^{t}+\bm{V}^{t}\big).
\end{align}

The parameters are then updated as
\begin{equation}
\bm{o}^{t+1}
\leftarrow
\bm{o}^{t}
-\eta_1
\left(\frac{\partial L_{\rho}}{\partial\bm{o}}\right)^t,
\qquad
\bm{\Theta}^{t+1}
\leftarrow
\bm{\Theta}^{t}
-\eta_2
\left(\frac{\partial L_{\rho}}{\partial\bm{\Theta}}\right)^t,
\label{eq:o_theta_update}
\end{equation}
where $\eta_1>0$ and $\eta_2>0$ are the learning rates for $\bm{o}$ and
$\bm{\Theta}$, respectively.

Since $L(\bm{o},\bm{\Theta})$ contains both $L_{\mathrm{RGB}}$ and
$L_{\mathrm{lang}}$, the language loss contributes directly to the opacity
update. Thus, before the following sparsification step selects the retained
Gaussians, their opacities have already been adapted using both RGB and semantic
supervision. The 3D positions and scene-level semantic dictionaries are
optimized by the task objective but receive no quantization penalty.

\paragraph{\Circled{2} Sparsification Update.}
In this step, the auxiliary variable $\bm{y}$ is updated by solving
$\min_{\bm{y}}g(\bm{y})+\tfrac{\rho_1}{2}
\|\bm{o}^{t+1}-\bm{y}+\bm{u}^{t}\|_2^2$.
The closed-form solution~\citep{parikh2014proximal,boyd2011distributed} is
\begin{align}
\bm{y}^{t+1}
&\leftarrow
\Pi_{\bm{\mathcal{S}}}
(\bm{o}^{t+1}+\bm{u}^{t}),
\label{eq:sparse}
\end{align}
where $\Pi_{\bm{\mathcal{S}}}$ denotes projection onto the
sparsity-constrained set. When
$\bm{\mathcal{S}}=\{\bm{y}\mid\mathbf{card}(\bm{y})\leq\kappa\}$, the
projection retains the $\kappa$ entries with the largest magnitudes and sets
all others to zero.

Because of the quadratic penalty term
$\tfrac{\rho_1}{2}\|\bm{o}^{t+1}-\bm{y}+\bm{u}^{t}\|_2^2$,
the projection is applied to $\bm{o}^{t+1}+\bm{u}^{t}$ rather than opacity
alone. The scaled dual variable $\bm{u}$ accumulates the discrepancy between
$\bm{o}$ and $\bm{y}$ across iterations. The nonzero entries of
$\bm{y}^{t+1}$ identify the retained Gaussians, and the same selection is
applied to their point-aligned RGB and semantic attributes.

\paragraph{\Circled{3} Vector Quantization Update.}
In the third step, we update the quantization state $\bm{\mathcal{Q}}$ and the
quantized Gaussian parameters $\bm{\hat{\Theta}}$. In implementation, this
operation is applied independently to different attribute groups using
separate codebooks; we suppress the group index for clarity.

The clusters and centroids in $\bm{\mathcal{Q}}$ are updated using standard
$k$-means. Each centroid is first computed as the mean of its assigned
parameter vectors, after which each vector is reassigned to its nearest
centroid. Formally, for $j=1,\dots,M$:
\begin{equation}
\resizebox{\linewidth}{!}{$%
\begin{aligned}
&\begin{cases}
\bm{q}_j^{t+1}
=
\dfrac{1}{|Q_j^t|}
\sum_{\bm{\theta}_i^{t+1}\in Q_j^t}
\bm{\theta}_i^{t+1}, \\[1mm]
Q_j^{t+1}
=
\Big\{
\bm{\theta}_i^{t+1}
\;\Big|\;
\operatorname*{arg\,min}_{k=1,\dots,M}
\|\bm{\theta}_i^{t+1}-\bm{q}_k^{t+1}\|^2=j
\Big\}.
\end{cases}
\end{aligned}
$}
\label{eq:quant_assign}
\end{equation}

We then project the dual-corrected Gaussian parameter vectors onto the updated
centroids:
\begin{align}
\bm{\hat{\Theta}}^{t+1}
&\leftarrow
\Pi_{\bm{\mathcal{Q}}^{t+1}}
(\bm{\Theta}^{t+1}+\bm{V}^{t}),
\label{eq:admm_projection}
\end{align}
where $\Pi_{\bm{\mathcal{Q}}}$ replaces each parameter vector with its nearest
cluster centroid.

Due to the quadratic penalty term
$\frac{\rho_2}{2}
\|\bm{\Theta}^{t+1}-\bm{\hat{\Theta}}+\bm{V}^{t}\|_2^2$,
the projection is applied to $\bm{\Theta}^{t+1}+\bm{V}^{t}$. The scaled dual
variable $\bm{V}$ tracks the discrepancy between $\bm{\Theta}$ and
$\bm{\hat{\Theta}}$ across iterations and carries this residual into
subsequent quantization updates.

\paragraph{\Circled{4} Dual Update.}
In the final step of iteration $t$, the scaled dual variables $\bm{u}$ and
$\bm{V}$ are updated as
\begin{align}
\bm{u}^{t+1}
&\leftarrow
\bm{u}^{t}+\bm{o}^{t+1}-\bm{y}^{t+1},
\label{eq:dual_u}\\
\bm{V}^{t+1}
&\leftarrow
\bm{V}^{t}+\bm{\Theta}^{t+1}-\bm{\hat{\Theta}}^{t+1}.
\label{eq:dual_V}
\end{align}

The update of $\bm{u}$ accumulates the residual between continuous opacity and
its sparse counterpart, while $\bm{V}$ performs the analogous correction for
the continuous and quantized attributes. These corrections enter the
sparsification and quantization operations of the next iteration.

The four updates alternate, allowing the trainable scene
parameters, retained Gaussian set, and quantized attributes to adapt under the
same RGB--semantic objective. We repeat this procedure until the residual
thresholds are met or the fixed iteration budget is exhausted. The nonzero
entries of the final $\bm{y}$ determine the retained Gaussians, while
$\bm{\hat{\Theta}}$ and the corresponding codebooks define their quantized
attributes. A convergence analysis of the resulting primal--dual scheme is
provided in
Sec.~\ref{app:convergence} of the supplementary material.

\section{Experiments}

\subsection{Settings}
\label{subsec:setup}

\paragraph{Benchmarks, metrics, and baselines.}
We evaluate on four LERF scenes~\citep{kerr2023lerf} and
ScanNet-10~\citep{dai2017scannet}, reporting semantic accuracy, RGB PSNR,
storage, and peak memory. Language baselines cover direct-lifting methods
(Occam's LGS and Dr.~Splat~\citep{cheng2025occam,jun2025dr}) and
render-supervised semantic fields (LangSplatV2, CCL-LGS, GAGS, and
CF3~\citep{li2025langsplatv2,tian2025ccl,peng2026gags,lee2025cf3}). Compression
baselines pair three Gaussian sparsifiers (opacity Top-$K$, PUP 3D-GS, and
GaussianSpa) with CompGS or C3DGS for attribute
quantization~\citep{hanson2025pup,zhang2025gaussianspa,navaneet2024compgs,niedermayr2024compressed}.

On \textit{LERF}, all query relevance maps use the same post-processing and a
fixed 0.4 threshold. mIoU averages query-mask IoU, while localization Acc.
checks whether the peak response lies inside a ground-truth box. On
\textit{ScanNet-10}, we follow OpenGaussian's label protocol and the ten-scene
split used by LUDVIG~\citep{wu2024opengaussian,marrie2025ludvig}. Each method
uses its native decoder and scale-selection rule to render ten class-score maps.
We bilinearly sample these scores at projected ScanNet points when the projected
and sensor depths differ by at most 5~cm, average them over valid views, and
take the pointwise argmax. Evaluation includes points observed in at least one
view after applying the same 0.1 activated-opacity threshold to each RGB 3DGS.
Per-scene
mIoU and mAcc average class IoU and recall over the ground-truth classes present
in that scene, followed by an unweighted average over the ten scenes.

\paragraph{Implementation details.}
Each scene starts from a 3DGS model~\citep{kerbl20233d} trained for 30k
iterations. We store all three LangSplatV2 semantic levels in one Gaussian
model and optimize it for 30k iterations; the released implementation instead
trains three separate level-specific models for 10k iterations
each~\citep{li2025langsplatv2}. The first 20k iterations follow the original
language-field training, while the final 10k activate our unified
sparsification--quantization updates. All rate-matched compressed variants use
the same 20\% Gaussian-retention budget. Detailed schedules are provided in the
supplement. Storage includes all files required for inference. Dr.~Splat
therefore includes its PQ index, whereas LangSplatV2 includes the semantic
coefficients and dictionaries for all three levels together with one shared
RGB 3DGS. Peak memory is measured from a fresh process through one
native-resolution text-query rendering.

\begin{table*}[!t]
    \centering
    \footnotesize
    \setlength{\tabcolsep}{1.25pt}
    \begin{minipage}[t]{0.49\textwidth}
        \centering
        \textbf{(a) LERF}
        \par\vspace{2pt}
\begin{tabular*}{\columnwidth}{@{\extracolsep{\fill}}lccccc@{}}
    \toprule
    Method & \multicolumn{3}{c}{Quality $\uparrow$} & \multicolumn{2}{c}{Efficiency $\downarrow$} \\
    \cmidrule(lr){2-4}\cmidrule(lr){5-6}
    & mIoU & Acc. & PSNR & \makecell{Storage\\(MB)} & \makecell{Peak Mem.\\(GB)} \\
    \midrule

    Occam's LGS & 56.8 & \rankthree{82.5} & 24.08 & 7663.23 & 17.83 \\

    Dr.\ Splat & 43.3 & 73.3 & \rankone{24.22} & 467.85 & 6.72 \\

    LangSplatV2 & \rankone{59.9} & \rankone{84.1} & \rankone{24.22} & \rankthree{345.94} & \rankthree{5.54} \\

    CCL-LGS & \rankthree{57.8} & 81.0 & \rankthree{24.11} & 735.93 & \ranktwo{5.21} \\

    GAGS & 53.9 & 75.7 & \rankone{24.22} & 399.06 & 8.96 \\

    CF3 & 54.2 & 74.2 & \rankone{24.22} & \ranktwo{327.91} & 8.29 \\

    \midrule

    \abbr{} & \ranktwo{58.3} & \ranktwo{83.0} & \ranktwo{24.15} & \rankone{59.35} & \rankone{4.60} \\

    \bottomrule
\end{tabular*}

    \end{minipage}\hfill
    \begin{minipage}[t]{0.49\textwidth}
        \centering
        \textbf{(b) ScanNet-10}
        \par\vspace{2pt}

\begin{tabular*}{\columnwidth}{@{\extracolsep{\fill}}lccccc@{}}
    \toprule
    Method & \multicolumn{3}{c}{Quality $\uparrow$} & \multicolumn{2}{c}{Efficiency $\downarrow$} \\
    \cmidrule(lr){2-4}\cmidrule(lr){5-6}
    & mIoU & Acc. & PSNR & \makecell{Storage\\(MB)} & \makecell{Peak Mem.\\(GB)} \\
    \midrule

    Occam's LGS & 19.8 & 36.0 & 20.73 & 595.00 & 3.75 \\

    Dr.\ Splat & \rankthree{45.5} & \rankthree{64.0} & \rankone{27.24} & 60.09 & \ranktwo{3.18} \\

    LangSplatV2 & \rankone{51.3} & \rankone{71.2} & \rankone{27.24} & \rankthree{42.90} & \rankthree{3.57} \\

    CCL-LGS & 25.4 & 40.3 & \rankthree{27.10} & 85.46 & 5.18 \\

    GAGS & 43.4 & 61.8 & \rankone{27.24} & 49.58 & 13.09 \\

    CF3 & 28.5 & 44.8 & \rankone{27.24} & \ranktwo{41.10} & 8.39 \\

    \midrule

    \abbr{} & \ranktwo{50.7} & \ranktwo{70.8} & \ranktwo{27.17} & \rankone{8.18} & \rankone{2.94} \\

    \bottomrule
\end{tabular*}

    \end{minipage}
    \caption{\textbf{Overall performance.}
    (a) LERF four-scene macro averages, where Acc. denotes localization
    accuracy. (b) ScanNet-10 ten-scene macro averages, where Acc. denotes mean
    class accuracy. \rankbox{rankone}{First}, \rankbox{ranktwo}{Second}, and
    \rankbox{rankthree}{Third}.}
    \label{tab:overall}
\end{table*}

\begin{figure*}[!t]
    \centering
    \includegraphics[width=\textwidth]{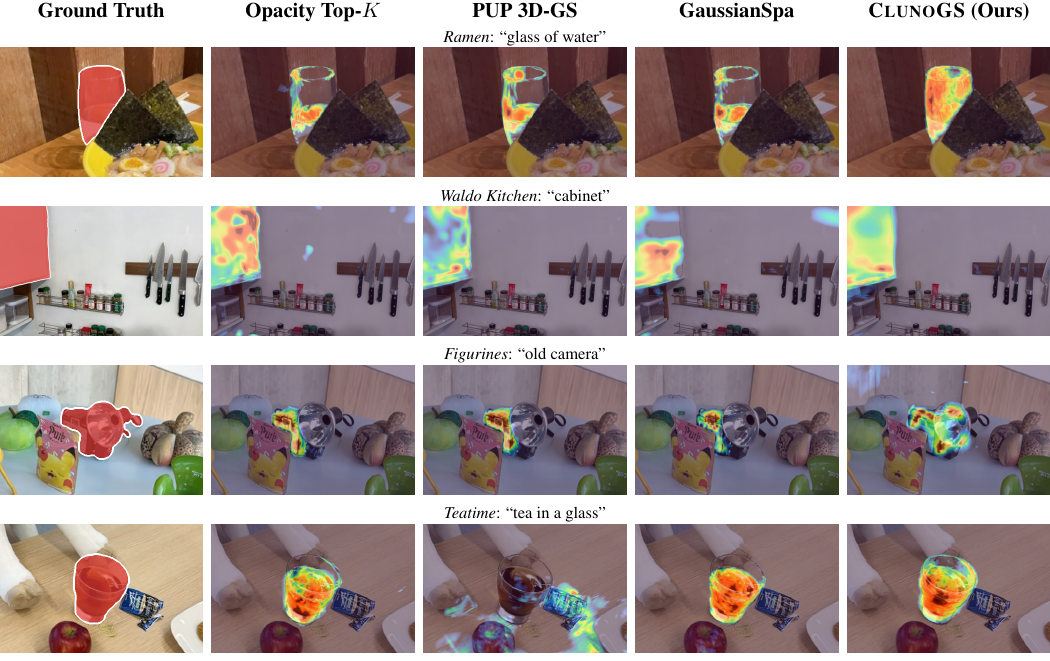}
    \caption{\textbf{PCA visualizations of rendered semantic maps.}
    Under the same 20\% Gaussian-retention budget, sequential compression
    blurs semantic boundaries and can remove semantic regions, whereas our
    joint method better preserves their structure.}
    \label{fig:compression_qualitative}
\end{figure*}

\subsection{Overall Performance}
\label{subsec:overall}

Table~\ref{tab:overall} evaluates whether aggressive compression can preserve
both semantic prediction accuracy and RGB rendering quality. Across LERF and
ScanNet-10,
we compare deployable systems from two Language 3DGS paradigms:
direct-lifting methods and render-supervised semantic fields. Semantic metrics
and PSNR measure quality, while storage and peak memory measure deployment cost.

Compared with CF3, the smallest competitor on both benchmarks, \abbr{} achieves
over $5\times$ storage reduction while also attaining the lowest peak memory.
It nevertheless ranks second in every semantic and RGB quality metric; its
maximum gaps to LangSplatV2 are only 1.6 mIoU points, 1.1 accuracy points, and
0.07~dB PSNR.
The same trend on both datasets shows that this advantage is not
benchmark-specific.

LangSplatV2 remains the quality leader because its full model retains detailed
semantic coefficients and dictionaries at all three scales. While
direct-lifting methods are designed for efficient querying, both Occam's LGS
and Dr.~Splat trail this render-supervised model in semantic quality. Occam's
LGS is also storage-intensive on both datasets and incurs the highest peak
memory on LERF.
\subsection{Sequential Compression Baselines}
\label{subsec:generic_compression}

Table~\ref{tab:compression_baselines} evaluates whether sequential
sparsification and quantization can match our joint formulation. All baselines
start from the same RGB 3DGS and retain 20\% of the Gaussians. For fairness,
optimization-based methods are embedded into the standard 30k-iteration
training without extra iterations. We pair three representative sparsifiers
(heuristic Opacity Top-$K$, importance-based PUP 3D-GS, and optimization-based
GaussianSpa) with CompGS or C3DGS, yielding six pipelines.

GaussianSpa~+~C3DGS achieves the highest sequential-baseline PSNR of 24.20~dB,
but all six pipelines degrade semantics, reaching at best 54.4 mIoU and 77.9
Acc. At 0.25M Gaussians and 59.35~MB, \abbr{} reaches 58.3 mIoU, 83.0 Acc.,
and 24.15~dB, improving the strongest sequential mIoU and Acc. by 3.9 and 5.1
points while remaining within 0.05~dB of the highest sequential PSNR. The PCA maps
in Figure~\ref{fig:compression_qualitative} show that sequential compression
blurs boundaries and can remove semantic regions, whereas ours better
preserves them.
\begin{table}[t]
    \centering
    \scriptsize
    \setlength{\tabcolsep}{0.7pt}
\begin{tabular*}{\columnwidth}{@{\extracolsep{\fill}}llccccc@{}}
    \toprule
    Sparsifier & Quantizer & \#G & mIoU $\uparrow$ & Acc. $\uparrow$ & PSNR $\uparrow$ & Storage $\downarrow$ \\
    \midrule

    None & None & 1.27M & 59.9 & 84.1 & 24.22 & 345.94 \\

    \midrule
    Opacity Top-$K$ & CompGS & 0.25M & 51.4 & 71.9 & 22.12 & 59.35 \\
    Opacity Top-$K$ & C3DGS & 0.25M & \rankthree{53.2} & \rankthree{77.5} & 22.01 & 64.26 \\
    \addlinespace[1.5pt]
    PUP 3D-GS & CompGS & 0.25M & 53.1 & 69.1 & 23.84 & 59.35 \\
    PUP 3D-GS & C3DGS & 0.25M & \ranktwo{54.4} & 73.5 & 23.97 & 64.26 \\
    \addlinespace[1.5pt]
    GaussianSpa & CompGS & 0.25M & 49.0 & 74.0 & \rankthree{24.01} & 59.35 \\
    GaussianSpa & C3DGS & 0.25M & 52.4 & \ranktwo{77.9} & \rankone{24.20} & 64.26 \\

    \midrule
    \multicolumn{2}{c}{\textbf{\abbr{}}} & 0.25M & \rankone{58.3} & \rankone{83.0} & \ranktwo{24.15} & 59.35 \\

    \bottomrule
\end{tabular*}

    \caption{\textbf{Sequential sparsification and quantization on LERF.}
    All compressed rows use the same 20\% Gaussian-retention budget; storage is
    in MB.
    Colors indicate quality rankings and exclude the uncompressed reference.}
    \label{tab:compression_baselines}
\end{table}
\subsection{Ordering Semantic Learning and Compression}
\label{subsec:ordered_pipelines}

Table~\ref{tab:ordered_pipelines} compares two sequential orders on LERF using
LangSplatV2 and Dr.~Splat. Original denotes the complete semantic model before
GaussianSpa~+~CompGS. For each method, both variants start from the same RGB
3DGS and use the same target pruning mask and compression configuration, with
the Gaussian count matched to ours.
The semantics-first order builds the native semantic representation before
compression and re-indexes the surviving records without retraining. The
compression-first order instead learns semantics directly on the compressed
RGB 3DGS. We retain each method's native representation rather than replacing
it with our VQ, and storage follows the complete-pipeline accounting of
Table~\ref{tab:overall}.

\begin{table}[t]
    \centering
    \scriptsize
    \setlength{\tabcolsep}{1.0pt}
\begin{tabular*}{\columnwidth}{@{\extracolsep{\fill}}llccccc@{}}
    \toprule
    Method & Pipeline & mIoU $\uparrow$ & Acc. $\uparrow$ & PSNR $\uparrow$ & \#G $\downarrow$ & \makecell{Storage\\(MB) $\downarrow$} \\
    \midrule

    \multirow{3}{*}{LangSplatV2} & Original & 59.9 & 84.1 & 24.22 & 1.27M & 345.94 \\

        & Sem. $\rightarrow$ Compress & 49.0 & 74.0 & 24.01 & 0.25M & \textbf{59.35} \\

        & Compress $\rightarrow$ Sem. & 52.4 & 77.9 & 23.80 & 0.25M & \textbf{59.35} \\

    \cmidrule(lr){1-7}
    \multirow{3}{*}{Dr.\ Splat} & Original & 43.3 & 73.3 & 24.22 & 1.27M & 467.85 \\

        & Sem. $\rightarrow$ Compress & 27.1 & 59.3 & 24.00 & 0.25M & 101.69 \\

        & Compress $\rightarrow$ Sem. & 29.1 & 60.8 & 24.00 & 0.25M & 101.69 \\

    \midrule
    \textbf{\abbr{}} & Joint & \textbf{58.3} & \textbf{83.0} & \textbf{24.15} & 0.25M & \textbf{59.35} \\

    \bottomrule
\end{tabular*}

    \caption{\textbf{Ordering semantic learning and compression.}}
    \label{tab:ordered_pipelines}
\end{table}

This comparison isolates the effect of ordering. Compression-first achieves
better semantic metrics for both methods, suggesting that learning semantics
on retained support is less damaging than pruning a completed semantic field.
However, both orders remain below their original models.
Our joint row is a system-level reference, while Table~\ref{tab:matched_rate_ablations}
provides controlled attribution with the backbone and \mbox{compression} operators
fixed. At the same 0.25M Gaussians and 59.35~MB of storage as the best
sequential semantic pipeline, \abbr{} improves mIoU from 52.4 to 58.3, Acc.
from 77.9 to 83.0, and PSNR from 23.80 to 24.15~dB. Thus, our joint
optimization mitigates the degradation from sequential compression and
semantics-agnostic Gaussian selection.

\section{Ablation Studies}
\label{sec:ablation}

\begin{table}[t]
    \centering
    \scriptsize
    \setlength{\tabcolsep}{1.5pt}
\begin{tabular*}{\columnwidth}{@{\extracolsep{\fill}}lccccc@{}}
    \toprule
    Variant & mIoU $\uparrow$ & Acc. $\uparrow$ & PSNR $\uparrow$ & \makecell{Storage\\(MB) $\downarrow$} & \#G $\downarrow$ \\
    \midrule
    Uncompressed model & 59.9 & 84.1 & 24.22 & 345.94 & 1.27M \\
    Sparsification only & 58.5 & 83.3 & 24.20 & 173.07 & 0.25M \\
    VQ only & 59.0 & 84.0 & 24.19 & 118.46 & 1.27M \\
    \midrule
    Full & 58.3 & 83.0 & 24.15 & 59.35 & 0.25M \\
    \bottomrule
\end{tabular*}

    \caption{\textbf{Component ablations on LERF.}
    The rows are not rate-matched; \#G and storage report each operating point.}
    \label{tab:compression_components}
\end{table}

Table~\ref{tab:compression_components} separates the two components:
sparsification reduces Gaussian support, whereas VQ compresses the retained
attributes. Sparsification removes 80.3\% of the Gaussians, while VQ reduces
storage to 34.2\% without pruning. Together they use only 17.2\% of the
original storage, with losses of 1.6 mIoU points, 1.1 accuracy points, and
0.07~dB. Removing VQ or sparsification increases storage by 2.9$\times$ or
2.0$\times$ for marginal quality gains, confirming their complementary
effects.

\begin{table}[t]
    \centering
    \scriptsize
    \setlength{\tabcolsep}{1.5pt}
\begin{tabular*}{\columnwidth}{@{\extracolsep{\fill}}llccc@{}}
    \toprule
    Group & Variant & mIoU $\uparrow$ & Acc. $\uparrow$ & PSNR $\uparrow$ \\
    \midrule
    \multirow{2}{*}{Design components} & w/o semantic guidance & 49.0 & 74.0 & 24.01 \\
    & Penalty only & 44.6 & 68.1 & 22.81 \\
    \cmidrule(lr){1-5}
    \multirow{3}{*}{Optimization strategy} & Sparsify $\rightarrow$ VQ & 48.3 & 70.0 & 22.35 \\
    & VQ $\rightarrow$ Sparsify & 47.2 & 68.9 & 23.24 \\
    & Joint (ours) & 58.3 & 83.0 & 24.15 \\
    \bottomrule
\end{tabular*}

    \caption{\textbf{Optimization ablations on LERF.}
    All variants use the same initialization, views, and budget, with 0.25M
    Gaussians and 59.35~MB of storage.}
    \label{tab:matched_rate_ablations}
\end{table}

Table~\ref{tab:matched_rate_ablations} isolates joint optimization at a fixed
rate. \mbox{Blocking} $L_{\mathrm{lang}}$ from opacity makes Gaussian selection
semantic-agnostic while retaining semantic learning and RGB supervision,
reducing mIoU and Acc. by 9.3 and 9.0 points with only a 0.14~dB PSNR loss.
The penalty-only variant keeps the penalties and projections but fixes
$\bm{u}=\bm{0}$ and $\bm{V}=\bm{0}$, removing residual correction. Its larger
drops of 13.7 mIoU points, 14.9 accuracy points, and 1.34~dB show that
primal--dual feedback is needed to coordinate the two projections. The
sequential variants apply the same sparsification and VQ operators in opposite
orders without joint co-adaptation. Joint optimization exceeds their best
value for each metric by 10.0 mIoU, 13.0 accuracy points, and 0.91~dB. Since
all variants use the same initialization, views, Gaussian count, and storage,
the gains come from semantics-aware selection and joint optimization rather
than the compression rate.

\FloatBarrier

\section{Conclusion}
We present \abbr{}, a unified framework that jointly sparsifies and quantizes
Language 3DGS under RGB and semantic supervision. Across LERF and ScanNet-10,
it substantially reduces storage and peak memory while retaining competitive
RGB and semantic quality. Controlled comparisons show that semantics-aware
selection and primal--dual co-adaptation mitigate the semantic degradation of
sequential compression.

\flushbottom
\clearpage
{
    \small
    \bibliographystyle{ieeenat_fullname}
    \bibliography{main}
}

\clearpage
\appendix
\setcounter{equation}{0}
\renewcommand{\theequation}{S\arabic{equation}}
\renewcommand{\theHequation}{supp.\arabic{equation}}
\setcounter{table}{0}
\renewcommand{\thetable}{S\arabic{table}}
\renewcommand{\theHtable}{supp.\arabic{table}}
\maketitlesupplementary
\section{Implementation Details}
\label{app:implementation}

\begin{table}[h]
    \centering
    \footnotesize
    \setlength{\tabcolsep}{3pt}
    \begin{tabular}{@{}p{0.48\columnwidth}>{\raggedright\arraybackslash}p{0.42\columnwidth}@{}}
        \toprule
        Setting & Value \\
        \midrule
        3DGS initialization & 30k iterations \\
        Semantic field & 3 levels \\
        Standard language-field stage & First 20k iterations \\
        Unified optimization stage & Final 10k iterations \\
        Gaussian-retention budget & 20\% \\
        SH codebook & 256 entries \\
        \bottomrule
    \end{tabular}
    \caption{\textbf{Main-model implementation settings.}}
    \label{tab:supp_impl_settings}
\end{table}

\paragraph{Training schedule.}
Each scene starts from a 3DGS model trained for 30k iterations, with all three
semantic levels optimized in the same Gaussian model. The first 20k iterations
follow standard language-field training; the final 10k activate sparsification,
vector quantization (VQ), and dual updates. Sparsity projection retains the $\kappa=0.2N$
largest-magnitude entries of the dual-corrected opacity vector.

\paragraph{Representation and export.}
The retention mask is shared by all point-aligned RGB and semantic attributes.
Separate codebooks quantize color and SH coefficients, scale, rotation, and
retained semantic coefficient values. Positions, opacity, and semantic
dictionaries remain in full precision; semantic atom indices are stored
separately. No additional learned decoder is introduced.

\section{Convergence Analysis and Proof}
\label{app:convergence}
\subsection{Preliminaries}
\label{sec:supp_preliminaries}
\noindent\textbf{Constrained Objective.}
Our optimization seeks to minimize an objective with sparsity and vector
quantization. By introducing auxiliary variables $\bm{y}$ and
$\bm{\hat{\Theta}}$ to decouple the non-differentiable terms, we reformulate
the problem into an equality-constrained form:
\begin{equation}
\begin{aligned}
\min_{\bm{o}, \bm{\Theta}, \bm{y}, \bm{\hat{\Theta}}, \bm{\mathcal{Q}}}
  L(\bm{o}, \bm{\Theta}) &+ g(\bm{y})
  + \sum_{j=1}^M \sum_{\bm{\hat{\theta}}_i\in Q_j}
  \|\bm{\hat{\theta}}_i - \bm{q}_j\|_2^2, \\[3pt]
\qquad \text{s.t. }&
  \bm{o} = \bm{y},
  \bm{\Theta} = \bm{\hat{\Theta}}.
\end{aligned}
\label{eq:supp_problem_formulation}
\end{equation}
Here, $\bm{\mathcal{Q}}$ contains both the cluster assignments and their
centroids and is optimized jointly with the other variables; it is not held
fixed in the analysis below.

\noindent\textbf{Augmented Lagrangian.}
To solve Eq.~\ref{eq:supp_problem_formulation}, we incorporate the constraints
through the augmented Lagrangian with scaled multipliers
\(\bm{u},\bm{V}\) and penalty coefficients \(\rho_1,\rho_2\):
\begin{equation}
\begin{aligned}
L_{\rho}(\bm{o}, \bm{\Theta}, &\bm{y}, \bm{\hat{\Theta}},\bm{u},\bm{V}, \bm{\mathcal{Q}}) = \\
& L(\bm{o}, \bm{\Theta}) + g(\bm{y}) + \sum_{j=1}^M \sum_{\bm{\hat{\theta}}_i\in Q_j}  \|\bm{\hat{\theta}}_i - \bm{q}_j\|_2^2 \\
&+ \frac{\rho_1}{2}\|\bm{o}-\bm{y}+\bm{u}\|_2^2
+ \frac{\rho_2}{2} \|\bm{\Theta} - \bm{\hat{\Theta}} + \bm{V}\|_2^2.
\end{aligned}
\label{eq:supp_Lagrangian}
\end{equation}

\noindent\textbf{Primal Updates.}
At iteration $t$, $\bm{o}^t$ and $\bm{\Theta}^t$ are updated by gradient
descent on the augmented Lagrangian:
\begin{equation}
\begin{aligned}
\bm{o}^{t+1}
&\leftarrow \bm{o}^t-\eta_1
\left(\frac{\partial L_{\rho}}{\partial\bm{o}}\right)^t,\\
\bm{\Theta}^{t+1}
&\leftarrow \bm{\Theta}^t-\eta_2
\left(\frac{\partial L_{\rho}}{\partial\bm{\Theta}}\right)^t,
\end{aligned}
\label{eq:supp_primal_updates}
\end{equation}
where $\eta_1, \eta_2$ are learning rates.

\noindent\textbf{Sparsification Update.}
The auxiliary variable $\bm{y}$ admits a closed-form update through the
sparsity operator $\Pi_{\bm{\mathcal{S}}}$:
\begin{align}
\bm{y}^{t+1} \leftarrow \Pi_{\bm{\mathcal{S}}}(\bm{o}^{t+1} + \bm{u}^t).
\label{eq:supp_sparse_update}
\end{align}
Concretely, $\Pi_{\bm{\mathcal{S}}}$ keeps the $\kappa$ entries of
$(\bm{o}^{t+1}+\bm{u}^t)$ with the largest magnitudes and zeros the rest.

\noindent\textbf{Quantization Update.}
For the quantization block, $\bm{\mathcal{Q}}$ is explicitly updated by the
two Lloyd steps used by standard $k$-means. For $j=1,\ldots,M$,
\begin{equation}
\begin{aligned}
\bm q_j^{t+1}
&=\frac{1}{|Q_j^t|}
\sum_{\bm\theta_i^{t+1}\in Q_j^t}\bm\theta_i^{t+1},\\
Q_j^{t+1}
&=\left\{\bm\theta_i^{t+1}:j\in
\operatorname*{arg\,min}_{k}\|\bm\theta_i^{t+1}-\bm q_k^{t+1}\|_2^2
\right\}.
\end{aligned}
\label{eq:supp_codebook_update}
\end{equation}
Ties are resolved deterministically. The quantized parameters are then
obtained by projecting the dual-corrected model parameters onto the updated
centroids:
\begin{equation}
\bm{\hat{\Theta}}^{t+1} \leftarrow
\Pi_{\bm{\mathcal{Q}}^{t+1}}(\bm{\Theta}^{t+1} + \bm{V}^t),
\label{eq:supp_quant_update}
\end{equation}
where $\Pi_{\bm{\mathcal{Q}}}$ maps each vector to its assigned centroid.

\subsection{Analysis and Proof}
\label{sec:supp_proof}
This section analyzes the proposed optimization scheme. We show that its
Lyapunov sequence converges and that every limit point satisfies the continuous
first-order conditions together with the projection and codebook fixed-point
conditions of Eq.~\ref{eq:supp_problem_formulation}. The proof retains the
standard sufficient-decrease structure for non-convex, non-smooth
primal--dual optimization.

\paragraph{Theoretical Assumptions.}
We use the following conditions:
\begin{enumerate}
    \item \textbf{Properties of the Loss Function.} \label{assump:smoothness}
    The loss $L(\bm{o},\bm{\Theta})$ is continuously differentiable and
    $L$-smooth in $(\bm{o},\bm{\Theta})$, i.e., for some $L>0$,
    \[
    \begin{aligned}
    &\|\nabla L(\bm{o}_1,\bm{\Theta}_1) - \nabla L(\bm{o}_2,\bm{\Theta}_2)\| \\
    &\le L\|(\bm{o}_1,\bm{\Theta}_1) - (\bm{o}_2,\bm{\Theta}_2)\|.
    \end{aligned}
    \]
    \item \textbf{Lower Boundedness.} \label{assump:lower_bound}
    The objective function, including the non-smooth terms, is bounded below
    by 0.
    \item \textbf{Penalty and Step Size.} \label{assump:stepsize}
    The penalty parameters $\rho_1, \rho_2 > 0$ and the learning rates
    $\eta_1, \eta_2 > 0$ satisfy
    \[
    \eta_1 < \frac{2}{L + \rho_1}, \quad
    \eta_2 < \frac{2}{L + \rho_2}.
    \]
    \item \textbf{Limit Point Existence.} \label{assump:limit}
    The sequence generated by our optimization scheme,
    $\{(\bm{o}_t,\bm{\Theta}_t,\bm{y}_t,\bm{\hat{\Theta}}_t,
    \bm{u}_t,\bm{V}_t,\bm{\mathcal{Q}}_t)\}_{t\in\mathbb{N}}$,
    has at least one limit point.
    \item \textbf{Quantization-Block Descent.} \label{assump:codebook}
    The $k$-means block has no empty clusters, uses deterministic tie breaking,
    and performs Lloyd updates until a coordinate-wise fixed point. Each
    centroid and assignment step therefore does not increase its within-cluster
    squared error. The resulting combined update of
    $(\bm{\mathcal{Q}},\bm{\hat{\Theta}})$ does not increase the corresponding
    augmented block objective.
    \item \textbf{Dual-residual Control.} \label{assump:dualcontrol}
    For the inexact primal updates, the dual-step contribution
    $\mathcal E_t$ defined below satisfies
    \[
    \mathcal E_t\le b_{\bm o}\|\Delta\bm o_t\|_2^2
    +b_{\bm\Theta}\|\Delta\bm\Theta_t\|_2^2
    -\sum_{i=1}^2 d_i\|\bm r_i^{t+1}\|_2^2,
    \]
    for $d_i>0$, $0\le b_{\bm o}<c_{\bm o}$, and
    $0\le b_{\bm\Theta}<c_{\bm\Theta}$. This is the relative-error condition
    required to keep the inexact primal and dual steps within the descent
    supplied by Lemma~1.
\end{enumerate}

\paragraph{Augmented Lagrangian and Lyapunov Function.}
For the augmented Lagrangian in Eq.~\ref{eq:supp_Lagrangian}, we use the
following residual-augmented Lyapunov function:
\begin{equation}
\begin{aligned}
\Phi_t={}&L_{\rho}(\bm{o}_t,\bm{\Theta}_t,\bm{y}_t,
\bm{\hat{\Theta}}_t,\bm{u}_t,\bm{V}_t,\bm{\mathcal{Q}}_t)\\
&+\frac{\gamma_1}{2}\|\bm{o}_t-\bm{y}_t\|_2^2
+\frac{\gamma_2}{2}\|\bm{\Theta}_t-\bm{\hat{\Theta}}_t\|_2^2,
\end{aligned}
\label{eq:supp_lyapunov}
\end{equation}
where $0<\gamma_i<\rho_i$. This definition makes explicit the residual
terms used in the descent accounting below.
For brevity, we introduce the notation:
\begin{equation}
\begin{aligned}
\Delta \bm{o}_t &= \bm{o}_{t+1} - \bm{o}_t,\\
\Delta \bm{\Theta}_t &= \bm{\Theta}_{t+1} - \bm{\Theta}_t,\\
\bm{r}^{ t+1}_{\bm{o}} &= \bm{o}_{t+1} - \bm{y}_{t+1},\\
\bm{r}^{ t+1}_{\bm{\Theta}} &= \bm{\Theta}_{t+1} - \bm{\hat{\Theta}}_{t+1}.
\end{aligned}
\label{eq:supp_residuals}
\end{equation}
The dual-step contribution in Assumption~\ref{assump:dualcontrol} is
\begin{equation}
\begin{aligned}
\mathcal E_t={}&\sum_{i=1}^{2}\frac{\rho_i}{2}
\left(2\langle\bm r_i^{t+1}+\bm u_i^t,\bm r_i^{t+1}\rangle
+\|\bm r_i^{t+1}\|_2^2\right)\\
&+\sum_{i=1}^{2}\frac{\gamma_i}{2}
\left(\|\bm r_i^{t+1}\|_2^2-\|\bm r_i^t\|_2^2\right).
\end{aligned}
\label{eq:supp_dual_control}
\end{equation}

\paragraph{Lemma 1 (Model-parameter update yields sufficient decrease).}
Let $\bm{o}_{t+1}$ and $\bm{\Theta}_{t+1}$ be updated as in
Eq.~\ref{eq:supp_primal_updates} under
Assumptions~\ref{assump:smoothness} and~\ref{assump:stepsize}. Then:
\begin{equation}
\begin{aligned}
    &L_{\rho}(\bm{o}_{t+1},\bm{\Theta}_{t+1},\bm{y}_t,\bm{\hat{\Theta}}_t,\bm{u}_t,\bm{V}_t,\bm{\mathcal{Q}}_t)\\
    &\le L_{\rho}(\bm{o}_t,\bm{\Theta}_t,\bm{y}_t,\bm{\hat{\Theta}}_t,\bm{u}_t,\bm{V}_t,\bm{\mathcal{Q}}_t) \\
    &\quad - c_{\bm{o}}\|\Delta \bm{o}_t\|_2^2 -c_{\bm{\Theta}}\|\Delta \bm{\Theta}_t\|_2^2,
\end{aligned}
\label{eq:supp_primal_decrease}
\end{equation}
with explicit positive constants
\begin{equation}
\begin{aligned}
    &c_{\bm{o}} = \frac{1}{\eta_1} - \frac{L}{2} - \frac{\rho_1}{2} > 0,\\
&c_{\bm{\Theta}} = \frac{1}{\eta_2} - \frac{L}{2} - \frac{\rho_2}{2} > 0.
\end{aligned}
\label{eq:supp_decrease_constants}
\end{equation}
\textbf{Proof.} With the auxiliary, dual, and codebook variables fixed, the
$L$-smoothness of $L$ and the two quadratic penalties bound the change across
the joint primal step as
\begin{equation}
\begin{aligned}
\Delta L_{\rho}
&\le \left\langle\nabla_{\bm o}L_{\rho},\Delta\bm o_t\right\rangle
+\frac{L+\rho_1}{2}\|\Delta\bm o_t\|_2^2\\
&\quad+\left\langle\nabla_{\bm\Theta}L_{\rho},
\Delta\bm\Theta_t\right\rangle
+\frac{L+\rho_2}{2}\|\Delta\bm\Theta_t\|_2^2.
\end{aligned}
\label{eq:supp_joint_descent}
\end{equation}
where both gradients are evaluated at iteration $t$. Substituting the two
gradient steps from
Eq.~\ref{eq:supp_primal_updates} yields
Eq.~\ref{eq:supp_primal_decrease} with the constants in
Eq.~\ref{eq:supp_decrease_constants}.

\paragraph{Lemma 2 (Auxiliary updates are non-increasing).}
The updates for $\bm{y}$, $\bm{\hat{\Theta}}$, and $\bm{\mathcal{Q}}$ do
not increase their corresponding block objectives. In particular:
\begin{equation}
\begin{aligned}
    &L_{\rho}(\bm{o}_{t+1},\bm{\Theta}_{t+1},\bm{y}_{t+1},\bm{\hat{\Theta}}_{t+1},\bm{u}_t,\bm{V}_t,\bm{\mathcal{Q}}_{t+1}) \\
    &\le L_{\rho}(\bm{o}_{t+1},\bm{\Theta}_{t+1},\bm{y}_t,\bm{\hat{\Theta}}_t,\bm{u}_t,\bm{V}_t,\bm{\mathcal{Q}}_t).
\end{aligned}
\label{eq:supp_auxiliary_decrease}
\end{equation}
\textbf{Proof.} The update for $\bm{y}_{t+1}$ in
Eq.~\ref{eq:supp_sparse_update} minimizes
$g(\bm{y})+\frac{\rho_1}{2}\|\bm{o}_{t+1}-\bm{y}+\bm{u}_t\|_2^2$.
For $\bm{\mathcal{Q}}$, the centroid update minimizes the within-cluster
squared error for fixed assignments, and the reassignment update minimizes it
for fixed centroids. Thus every Lloyd step in
Eq.~\ref{eq:supp_codebook_update} is non-increasing; no global $k$-means
optimum is assumed. Assumption~\ref{assump:codebook} further requires the
combined quantization update, including the nearest-centroid projection in
Eq.~\ref{eq:supp_quant_update}, to be non-increasing in its augmented block
objective. Chaining the auxiliary updates proves the stated inequality.

\paragraph{Lemma 3 (Dual ascent and Lyapunov accounting).}
Let $\Phi_{t+1/2}$ denote the Lyapunov value after the primal, auxiliary, and
codebook updates but before the dual update. Writing $\bm{u}^1=\bm{u}$ and
$\bm{u}^2=\bm{V}$, we have
\begin{equation}
\begin{aligned}
    &\bm{u}^1_{t+1} = \bm{u}^1_t + (\bm{o}_{t+1} - \bm{y}_{t+1}) = \bm{u}^1_t + \bm{r}^{t+1}_1,\\
&\bm{u}^2_{t+1} = \bm{u}^2_t + (\bm{\Theta}_{t+1} - \bm{\hat{\Theta}}_{t+1}) = \bm{u}^2_t + \bm{r}^{ t+1}_2,
\end{aligned}
\label{eq:supp_dual_updates}
\end{equation}
where $\bm{r}_1:=\bm{r}_{\bm{o}}$ and
$\bm{r}_2:=\bm{r}_{\bm{\Theta}}$. The augmented-Lagrangian change across
the dual step is:
\begin{equation}
\begin{aligned}
&L_{\rho}(\bm{o}_{t+1},\bm{\Theta}_{t+1},\bm{y}_{t+1},\bm{\hat{\Theta}}_{t+1},\bm{u}^1_{t+1},\bm{u}^2_{t+1},\bm{\mathcal{Q}}_{t+1})\\
&\quad - L_{\rho}(\bm{o}_{t+1},\bm{\Theta}_{t+1},\bm{y}_{t+1},\bm{\hat{\Theta}}_{t+1},\bm{u}^1_t,\bm{u}^2_t,\bm{\mathcal{Q}}_{t+1}) \\
&= \sum_{i=1}^{2} \frac{\rho_i}{2}\Big( 2\langle \bm{r}^{ t+1}_i + \bm{u}^t_i, \bm{r}^{t+1}_i \rangle + \|\bm{r}^{t+1}_i\|_2^2 \Big),
\end{aligned}
\label{eq:supp_dual_change}
\end{equation}
and consequently:
\begin{equation}
\Phi_{t+1} = \Phi_{t+1/2} + \sum_{i=1}^{2} \frac{\rho_i}{2}\Big( 2\langle\bm{r}^{ t+1}_i + \bm{u}^t_i,\bm{r}^{ t+1}_i \rangle + \|\bm{r}^{ t+1}_i\|_2^2 \Big),
\label{eq:supp_dual_lyapunov}
\end{equation}
because the two residual terms in $\Phi$ do not change during the dual step.

\textbf{Proof.} Only the quadratic penalty involving $\bm{u}_i$ changes.
For $i\in\{1,2\}$, with $\bm{r}^{t+1}_i$ fixed and
$\bm{u}^i_{t+1}=\bm{u}^i_t+\bm{r}^{t+1}_i$, we have:
\begin{equation}
\begin{aligned}
    &\frac{\rho_i}{2}\Big(\|\bm{r}^{ t+1}_i + \bm{u}^i_{t+1}\|_2^2 - \|\bm{r}^{ t+1}_i + \bm{u}^i_t\|_2^2\Big) =\\
    &\frac{\rho_i}{2}\Big( 2\langle \bm{r}^{ t+1}_i + \bm{u}^i_t, \bm{r}^{ t+1}_i \rangle + \|\bm{r}^{ t+1}_i\|_2^2 \Big).
\end{aligned}
\label{eq:supp_dual_identity}
\end{equation}
This follows from $\|\bm p+\bm q\|_2^2-\|\bm p\|_2^2
=2\langle\bm p,\bm q\rangle+\|\bm q\|_2^2$. Summing over both blocks
gives Eq.~\ref{eq:supp_dual_change}; the unchanged residual terms then give
Eq.~\ref{eq:supp_dual_lyapunov}.

\paragraph{Lemma 4 (Monotonicity, summability, and vanishing residuals).}
\label{lem:lem4}
Under Lemmas 1--3 and the standing assumptions, there are positive constants
$C_1,C_2$ such that
\begin{equation}
\Phi_{t+1} \le \Phi_t - C_1 \|\Delta \bm{o}_t\|_2^2 - C_2\|\Delta \bm{\Theta}_t\|_2^2,
\label{eq:supp_lyapunov_descent}
\end{equation}
and hence $\{\Phi_t\}$ is monotonically non-increasing and converges to a
finite limit $\Phi^*$. Moreover,
\begin{equation}
\begin{aligned}
    &\sum_{t=0}^{\infty}\Big(\|\Delta \bm{o}_t\|_2^2 + \|\Delta \bm{\Theta}_t\|_2^2\Big) < \infty,\\
&\lim_{t\to\infty}\|\bm{o}_t - \bm{y}_t\|_2 = 0,\\
&\lim_{t\to\infty}\|\bm{\Theta}_t - \bm{\hat{\Theta}}_t\|_2 = 0.
\end{aligned}
\label{eq:supp_summability}
\end{equation}
\textbf{Proof.} From Lemma 1 we have
\begin{equation}
\begin{aligned}
    &L_{\rho}(\bm{o}_{t+1}, \bm{\Theta}_{t+1}, \bm{y}_t, \bm{\hat{\Theta}}_t, \bm{u}^t_1, \bm{u}^t_2, \bm{\mathcal{Q}}_t)\\
    &\le L_{\rho}(\bm{o}_t, \bm{\Theta}_t, \bm{y}_t, \bm{\hat{\Theta}}_t, \bm{u}^t_1, \bm{u}^t_2, \bm{\mathcal{Q}}_t) \\
    &\quad - c_{\bm{o}} \|\Delta \bm{o}_t\|_2^2 - c_{\bm{\Theta}} \|\Delta \bm{\Theta}_t\|_2^2,
\end{aligned}
\label{eq:supp_primal_bound}
\end{equation}
and by Lemma~2, applying the non-increasing property of the
$\bm{y}$ and $\bm{\hat{\Theta}}$ updates, we obtain
\begin{equation}
\begin{aligned}
    &L_{\rho}(\bm{o}_{t+1}, \bm{\Theta}_{t+1}, \bm{y}_{t+1}, \bm{\hat{\Theta}}_{t+1}, \bm{u}^t_1, \bm{u}^t_2, \bm{\mathcal{Q}}_{t+1}) \\
    &\le L_{\rho}(\bm{o}_{t+1}, \bm{\Theta}_{t+1}, \bm{y}_t, \bm{\hat{\Theta}}_t, \bm{u}^t_1, \bm{u}^t_2, \bm{\mathcal{Q}}_{t}).
\end{aligned}
\label{eq:supp_auxiliary_bound}
\end{equation}
Combining Eqs.~\ref{eq:supp_primal_bound}
and~\ref{eq:supp_auxiliary_bound} and then applying Lemma~3
(Eq.~\ref{eq:supp_dual_lyapunov}) gives
\begin{equation}
\begin{aligned}
    &\Phi_{t+1} \le \Phi_t - c_{\bm{o}} \|\Delta \bm{o}_t\|_2^2 - c_{\bm{\Theta}} \|\Delta \bm{\Theta}_t\|_2^2 \\
    &\quad + \sum_{i=1}^{2} \frac{\rho_i}{2}\Big( 2 \langle \bm{r}^{ t+1}_i + \bm{u}^t_i, \bm{r}^{ t+1}_i \rangle + \| \bm{r}^{ t+1}_i \|_2^2 \Big)\\
    &\quad +\sum_{i=1}^{2}\frac{\gamma_i}{2}
    \left(\|\bm r_i^{t+1}\|_2^2-\|\bm r_i^t\|_2^2\right).
\end{aligned}
\label{eq:supp_combined_descent}
\end{equation}
The last two lines of Eq.~\ref{eq:supp_combined_descent} are exactly
$\mathcal E_t$ in
Eq.~\ref{eq:supp_dual_control}. Applying
Assumption~\ref{assump:dualcontrol} and setting
$C_1=c_{\bm o}-b_{\bm o}>0$ and
$C_2=c_{\bm\Theta}-b_{\bm\Theta}>0$ gives
\begin{equation}
\Phi_{t+1}\le\Phi_t-C_1\|\Delta\bm o_t\|_2^2
-C_2\|\Delta\bm\Theta_t\|_2^2
-\sum_{i=1}^2d_i\|\bm r_i^{t+1}\|_2^2.
\label{eq:supp_strict_descent}
\end{equation}
Since $\Phi_t$ is bounded below, summing
Eq.~\ref{eq:supp_strict_descent} over $t$ proves both the square summability of
the model-parameter differences and the vanishing of the two primal residuals.

\par\medskip
\noindent\textbf{Theorem 1 (Limit-point Stationarity).}\par
\noindent Suppose the standing assumptions hold for the sequence generated by
our optimization scheme. Then:
\begin{enumerate}
    \item\label{item:item1} The sequence $\{\Phi_t\}$ is non-increasing and
    converges to a finite limit $\Phi^*$.
    \item\label{item:item2} Both successive model-parameter differences vanish:
    $\lim_{t\to\infty}\|\bm{o}_{t+1}-\bm{o}_t\|_2=0$ and
    $\lim_{t\to\infty}\|\bm{\Theta}_{t+1}-\bm{\Theta}_t\|_2=0$.
    \item\label{item:item3} Both primal residuals vanish:
    $\lim_{t\to\infty}\|\bm{o}_t-\bm{y}_t\|_2=0$ and
    $\lim_{t\to\infty}\|\bm{\Theta}_t-\bm{\hat{\Theta}}_t\|_2=0$.
    \item\label{item:item4} Any limit point satisfies the continuous KKT
    conditions and the discrete projection fixed-point conditions of
    Eq.~\ref{eq:supp_problem_formulation}. Specifically:
\begin{itemize}
    \item \textbf{Stationarity for Model Parameters:} \\
    $\nabla_{\bm{o}} L(\bm{o}^*,\bm{\Theta}^*) + \rho_1 \bm{u}^* = \bm{0}$,\\ $\nabla_{\bm{\Theta}} L(\bm{o}^*,\bm{\Theta}^*) + \rho_2 \bm{V}^* = \bm{0}$.

    \item \textbf{Optimality for Auxiliary Variables:} \\
    $\bm{0} \in \partial g(\bm{y}^*) - \rho_1\big(\bm{o}^* - \bm{y}^* + \bm{u}^*\big)$, \\
    $\bm{\hat{\Theta}}^* = \Pi_{\bm{\mathcal{Q}}^*}\!\big(\bm{\Theta}^* + \bm{V}^*\big)$.

    \item \textbf{Optimality for the Codebook and Assignments:} for every
    nonempty $Q_j^*$,
    \begin{align*}
    \bm q_j^*&=\frac{1}{|Q_j^*|}
    \sum_{\bm\theta_i^*\in Q_j^*}\bm\theta_i^*,\\
    Q_j^*&=\left\{\bm\theta_i^*:j\in
    \operatorname*{arg\,min}_{k=1,\ldots,M}
    \|\bm\theta_i^*-\bm q_k^*\|_2^2\right\}.
    \end{align*}

    \item \textbf{Primal Feasibility:}
    $\bm{o}^* = \bm{y}^*$, $\bm{\Theta}^* = \bm{\hat{\Theta}}^*$.
\end{itemize}
\end{enumerate}
\textbf{Proof.} Items 1--3 follow from Lemma~4. For Item 4, let
$(\bm{o}^{*},\bm{\Theta}^{*},\bm{y}^{*},\bm{\hat{\Theta}}^{*},
\bm{u}^{*},\bm{V}^{*},\bm{\mathcal{Q}}^{*})$ be the limit of a subsequence
$\{(\bm{o}_{t_j},\bm{\Theta}_{t_j},\bm{y}_{t_j},
\bm{\hat{\Theta}}_{t_j},\bm{u}_{t_j},\bm{V}_{t_j},
\bm{\mathcal{Q}}_{t_j})\}_{j\in\mathbb{N}}$. The opacity update is
\[
\bm{o}_{t_j+1} \leftarrow \bm{o}_{t_j}-\eta_1\Big(\nabla_{\bm{o}} L(\bm{o}_{t_j},\bm{\Theta}_{t_j}) + \rho_1\big(\bm{o}_{t_j} - \bm{y}_{t_j} + \bm{u}_{t_j}\big)\Big).
\]
Item~2 makes the update direction vanish along this subsequence. Taking the
limit and using Item~3 gives stationarity for $\bm{o}^*$; the same argument applies
to $\bm{\Theta}^*$. The auxiliary-variable conditions and primal feasibility
follow from their update rules and Item 3. Under
Assumption~\ref{assump:codebook}, the centroid and assignment maps in
Eq.~\ref{eq:supp_codebook_update} are fixed at the limit, giving the stated
condition for $\bm{\mathcal{Q}}^*$. Thus, the limit point satisfies all stated
continuous and discrete stationarity conditions.

\end{document}